\documentclass[runningheads]{llncs}

\usepackage[T1]{fontenc}
\usepackage{graphicx}
\usepackage{amsmath}
\usepackage{xcolor}
\usepackage{amssymb}
\usepackage[linesnumbered,lined,ruled,noend]{algorithm2e}
\usepackage{booktabs}
\usepackage{multirow}
\usepackage{wrapfig}

\begin{document}

\title{Win-k: Improved Membership Inference Attacks on Small Language Models}

\titlerunning{Win-k: Improved MIAs on Small Language Models}

\author{Roya Arkhmammadova, Hosein Madadi Tamar, and M. Emre Gursoy}

\authorrunning{R. Arkhmammadova et al.}

\institute{Department of Computer Engineering, Koç University, Istanbul, Turkey \email{\{rarkhmammadova22, htamar24, emregursoy\}@ku.edu.tr}}

\maketitle  

\begin{abstract}
Small language models (SLMs) are increasingly valued for their efficiency and deployability in resource-constrained environments, making them useful for on-device, privacy-sensitive, and edge computing applications. On the other hand, membership inference attacks (MIAs), which aim to determine whether a given sample was used in a model’s training, are an important threat with serious privacy and intellectual property implications. In this paper, we study MIAs on SLMs. Although MIAs were shown to be effective on large language models (LLMs), they are relatively less studied on emerging SLMs, and furthermore, their effectiveness decreases as models get smaller. Motivated by this finding, we propose a new MIA called win-k, which builds on top of a state-of-the-art attack (min-k). We experimentally evaluate win-k by comparing it with five existing MIAs using three datasets and eight SLMs. Results show that win-k outperforms existing MIAs in terms of AUROC, TPR @ 1\% FPR, and FPR @ 99\% TPR metrics, especially on smaller models. 

\keywords{Small language models \and membership inference attacks \and privacy \and AI security \and responsible AI}
\end{abstract}

\vspace{-8pt}
\section{Introduction} \label{sec:Introduction}
\vspace{-4pt}

Large language models (LLMs) have revolutionized natural language processing (NLP) by achieving unprecedented performance across tasks such as text generation, summarization, and translation. However, the growing demand for resource-efficient NLP solutions has catalyzed a shift towards small language models (SLMs), which offer a lightweight yet effective alternative \cite{biderman2023pythia,hu2024minicpm,liu2024mobilellm}. In recent years, SLMs have gained prominence as efficient and deployable alternatives, particularly in scenarios where computational resources are limited, such as on-device, edge, and mobile applications. Techniques such as pruning, quantization, and knowledge distillation enable SLMs to retain core NLP capabilities while substantially reducing memory and latency requirements \cite{lu2024small,van2024survey}. 



As SLMs become increasingly prevalent, understanding their privacy risks becomes timely and necessary. A prominent risk is membership inference attacks (MIAs), where an adversary aims to determine whether a given data sample was used during a model's training process \cite{shokri2017membership,truex2019demystifying}. While MIAs have been studied in the context of large language models (LLMs) \cite{carlini2021extracting,duan2024membership,mattern2023membership,shi2024detecting}, their applicability and effectiveness in SLMs remain underexplored. Yet, MIAs have serious privacy and intellectual property implications in SLMs similar to LLMs.

In this paper, we focus on the application of MIAs on SLMs. First, we identify five popular MIAs in the realm of LLMs (loss, lowercase, zlib, neighborhood, and min-k) and execute them on three SLM families containing models with varying sizes: GPT-Neo, Pythia, and MobileLLM. Upon experimenting with multiple datasets, we identify a clear trend: As model sizes get smaller, the effectiveness of existing MIAs decreases. This key observation motivated us to propose a new MIA that is more effective, specifically in the context of SLMs. We therefore propose win-k, a new MIA which builds on top of a state-of-the-art attack, min-k \cite{shi2024detecting}. Min-k is a token-level attack which takes into account the bottom k\% fraction of token-level log probabilities when constructing a sample's membership score. In contrast, win-k proposes to compute window-level scores rather than token-level scores. That is, win-k slides over windows of consecutive tokens to compute their average log probability, and then uses the bottom k\% fraction of scores to construct the membership score. This approach helps in reducing the high variance which may be present in individual tokens' log probabilities, but cancels out when a window of consecutive tokens is considered.

We experimentally evaluate win-k by comparing it with five MIAs using three datasets and eight SLMs. Three metrics are used in the experimental comparisons: AUROC which captures the general effectiveness of MIAs across varying membership score thresholds; and TPR @ 1\% FPR and FPR @ 99\% TPR metrics which capture the behavior of MIAs under strict TPR or FPR requirements. Results show that win-k outperforms existing attacks in a large majority of cases. Results also show that win-k performs particularly better than other MIAs when model sizes are smaller. Through hyperparameter analyses, we offer insights into how the window size parameter $w$ and the fraction parameter $k$ should be selected in win-k in order to improve attack effectiveness. 

\textbf{Contributions.} In summary, our main contributions include:
\vspace{-3pt}
\begin{itemize}
\item We initiate the study of membership inference attacks (MIAs) on small language models (SLMs). We empirically demonstrate that although MIAs are effective on SLMs, their effectiveness declines as model size decreases.
\item Motivated by this finding, we propose a new MIA called win-k, which extends the state-of-the-art min-k attack by computing log probability scores over sliding windows of consecutive tokens, thereby mitigating the noise and outlier sensitivity observed in token-level analyses on small models.
\item We show that win-k outperforms existing MIAs through comprehensive experiments involving three datasets, eight SLMs, and three metrics. Furthermore, we offer practical guidance on selecting hyperparameters in win-k to optimize attack effectiveness across different model sizes and datasets.
\end{itemize}


\vspace{-8pt}
\section{Background and Preliminaries} \label{sec:Background}

\vspace{-4pt}
\subsection{Language Models}

In the last few years, language models have revolutionized the research in natural language processing (NLP) \cite{brown2020language,chang2024survey,liang2022holistic}. Say that we are given a vocabulary $\mathcal{V}$. A textual sample $x$ consists of a sequence of tokens: $x = (x_1,x_2,...,x_T)$ where each token $x_t \in \mathcal{V}$. Given $\mathcal{V}$, the objective of a language model is to maximize the likelihood of observed sequences, which can be expressed using the chain rule:
\begin{equation}
Pr(x_1, x_2, \dots, x_T) = \prod_{t=1}^{T} Pr(x_t \mid x_{<t})
\end{equation}
where $x_{<t} = (x_1, x_2, \dots, x_{t-1})$ denotes the preceding context. This decomposition enables language models to sequentially predict each token conditioned on prior context.

Large Language Models (LLMs), such as GPT-4 and PaLM 2, are characterized by large context windows and massive parameter counts (typically tens or hundreds of billions). Such massive parameter counts cause computational challenges concerning storage, training, and inference \cite{biderman2023pythia,brown2020language}. In contrast, Small Language Models (SLMs) are lightweight and designed for efficient deployment in resource-constrained settings. They typically have hundreds of millions or a few billion parameters, and therefore they are at least an order of magnitude smaller than LLMs \cite{biderman2023pythia,hu2024minicpm,liu2024mobilellm}. In this paper, we focus on SLMs. 

\vspace{-3pt}
\subsection{Membership Inference Attacks} \label{sec:MIA}

Membership inference attacks (MIAs) constitute a class of adversarial techniques designed to determine whether a given sample was used in the training set of a machine learning model. While MIAs were originally proposed in the context of classification models \cite{shokri2017membership,truex2019demystifying}, they are recently being adapted and applied to the context of LLMs \cite{duan2024membership,mattern2023membership,shi2024detecting}.

Let $\mathcal{M}$ denote a language model and $\mathcal{L}(x;M)$ denote the loss of sample $x$ on model $\mathcal{M}$. A MIA constructs a membership score $f(x;M)$ that is used to predict whether $x$ was a member $\mathcal{M}$'s training data. The membership score $f(x;M)$ is then compared against a threshold (say $\delta$) to predict $x$'s membership. The construction of $f(x;M)$ may often utilize $\mathcal{L}(x;M)$, but it differs from one MIA to another. Below, we introduce five state-of-the-art MIAs against LLMs and explain how they construct $f(x;M)$.

\textbf{Loss.} The Loss attack \cite{carlini2019secret,yeom2018privacy} is predicated on the observation that a model typically yields lower loss values for samples encountered during training. It simply uses the value of $\mathcal{L}$ as the membership score:
\begin{equation}
    f(x;M) = \mathcal{L}(x;M)
\end{equation}

\textbf{Lowercase.} The Lowercase attack \cite{carlini2021extracting} takes advantage of the sensitivity of language models to case-specific features. It converts the original sample to its lowercase version. After that, it compares the model's losses between the original and lowercase versions. If the model has substantially different loss values for the original sample and its lowercase version, it is likely that the case-specific features of the sample were memorized during training.
\begin{equation}
    f(x;M) = \frac{\mathcal{L}(\text{lowercase}(x);M)}{\mathcal{L}(x;M)}
\end{equation}

\textbf{Zlib} \cite{carlini2021extracting} employs $\mathcal{L}(x;M)$ together with the size of the compressed version of the sample using zlib compression. Let zlib($x$) denote the length in bytes of the zlib compressed version of $x$. Then:
\begin{equation}
    f(x;M) = \frac{\mathcal{L}(x;M)}{\text{zlib}(x)}
\end{equation}

\textbf{Neighborhood} attack \cite{mattern2023membership} is based on the principle that models exhibit lower loss on memorized training samples compared to similar but non-training samples. It generates a set of synthetic neighbor texts for a given sample using a masked language model by applying word substitutions. Then, the attack compares the model’s loss on the original sample to the average loss across its synthetically generated neighbors. If the prior is significantly lower than the latter, the attack infers that the sample was likely present in the training data. Formally, for an input sample $x$ and its $n$ generated neighbors $\{ \tilde{x}^1, \tilde{x}^2, \dots, \tilde{x}^n \}$:
\begin{equation}
    f(x;M) = \mathcal{L}(x;M) - \frac{1}{n} \sum\limits_{i=1}^{n} \mathcal{L}(\tilde{x}^i; M)
\end{equation}

\textbf{Min-k} \cite{shi2024detecting} is based on the hypothesis that non-member samples are more likely to include a few outlier words with low log-likelihood (i.e., low probability), while a member sample is less likely to do so. Given sample $x = (x_1,x_2,...,x_T)$ and hyperparameter $k$, let $\text{min-k}(x)$ denote the set formed by the $k\%$ of tokens in $x$ with minimum probability. Then:
\begin{equation}
    f(x;M) = \frac{1}{|\text{min-k}(x)|} \sum\limits_{x_i \in \text{min-k}(x)} \text{log}(Pr(x_i|x_{<i}))
\end{equation}

\subsection{How Do MIAs Perform on SLMs?}

\begin{figure*}[!t]
    \centering
    \includegraphics[width=.3\linewidth]{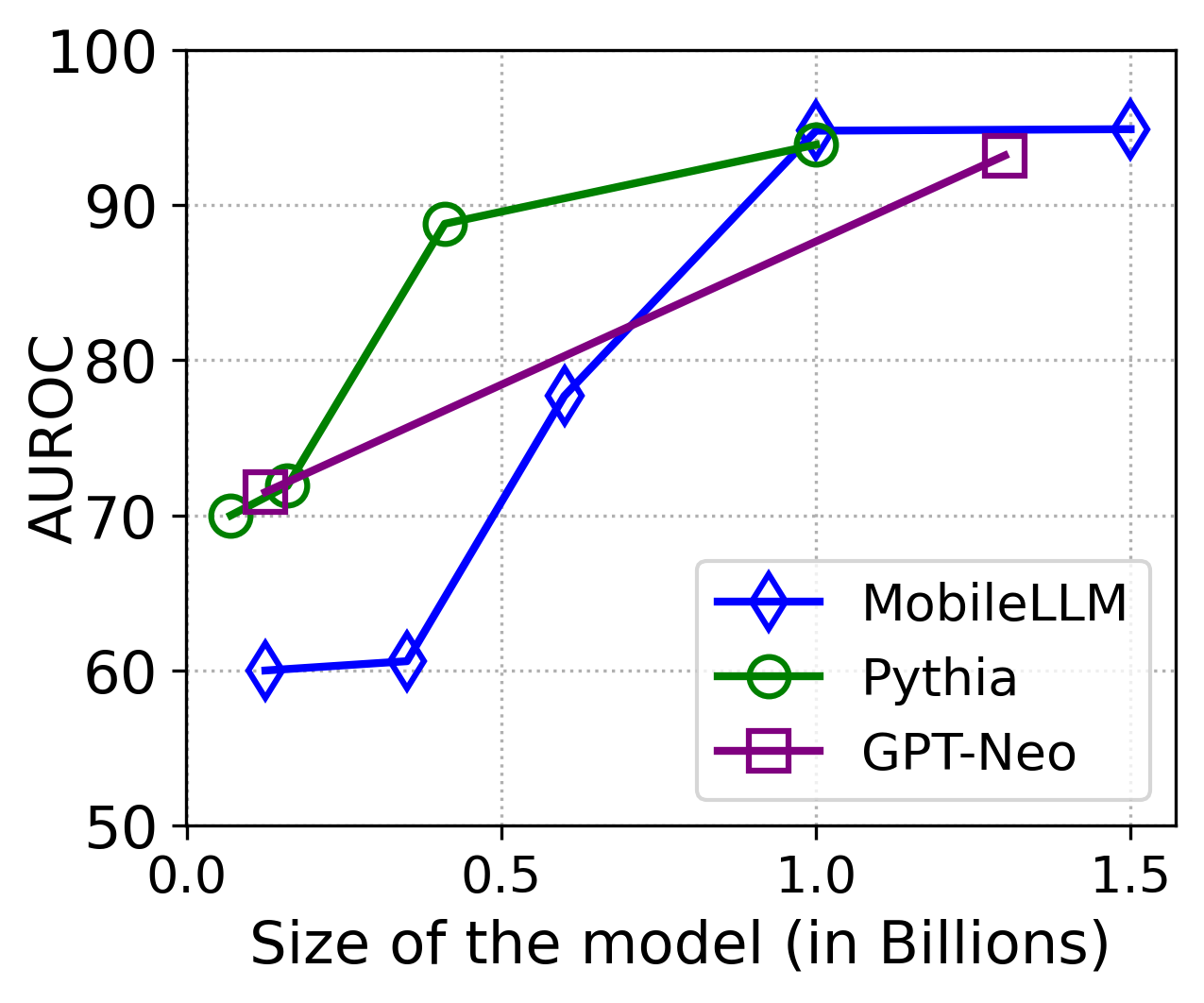} \hspace{2pt}
    \includegraphics[width=.3\linewidth]{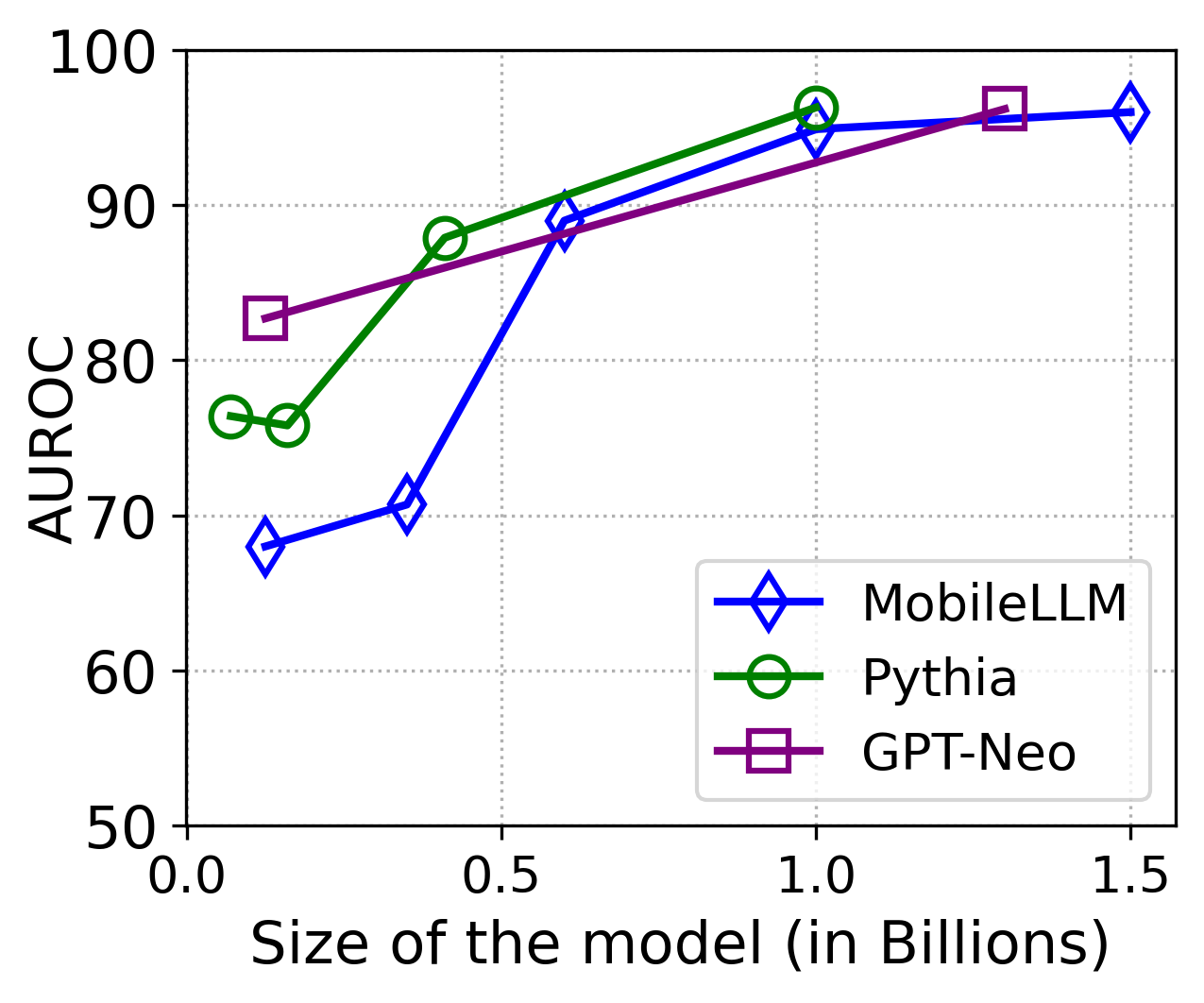} \hspace{2pt}
    \includegraphics[width=.3\linewidth]{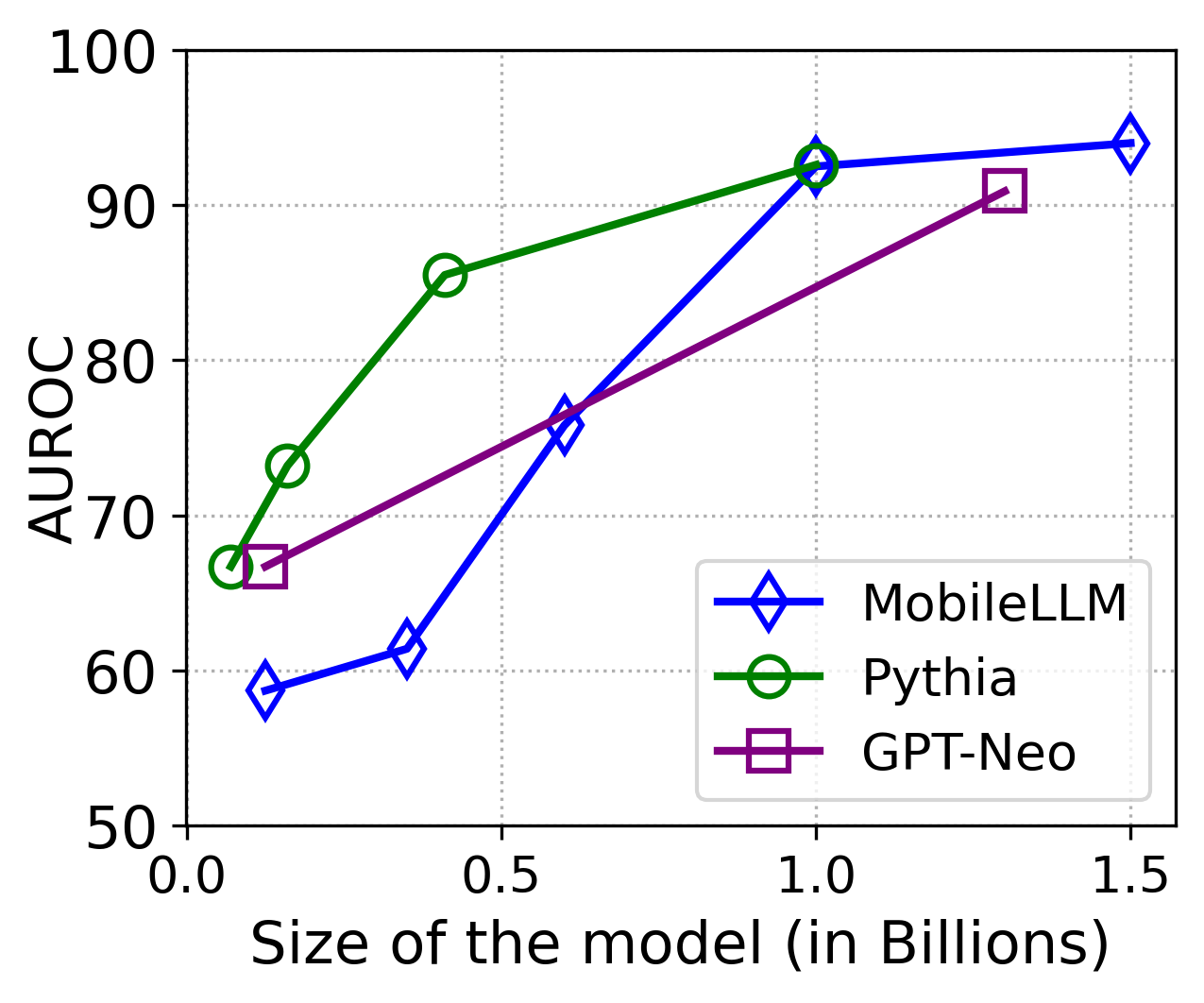}
    \vspace{-6pt}
    \caption{Average AUROCs of the five MIAs on SLMs with varying sizes (left plot: WikiText dataset, middle plot: AGNews dataset, right plot: XSum dataset).}
    \vspace{-6pt}
    \label{fig:MIAonSLM}
\end{figure*}

Previous literature has shown that MIAs are effective on LLMs \cite{carlini2021extracting,mattern2023membership,shi2024detecting}. In this paper, we focus on the applications of MIAs to SLMs. First, we measure the effectiveness of MIAs on SLMs with varying model sizes, i.e., varying number of parameters. To perform this experiment, we identified three model families which contain SLMs with varying sizes: GPT-Neo \cite{gpt-neo}, Pythia \cite{biderman2023pythia}, and MobileLLM \cite{liu2024mobilellm}. We fine-tuned these SLMs using three datasets that are commonly used in the literature: WikiText, AGNews, and XSum. (More details regarding the models, datasets, and the fine-tuning process can be found in Section \ref{sec:ExperimentSetup}.) We executed each of the five MIAs from Section \ref{sec:MIA} and measured their average AUROC scores (AUROC is a metric commonly used in the MIA literature to measure attack effectiveness).

The results of this experiment are shown in Figure \ref{fig:MIAonSLM} for each of the three datasets separately. The sizes of the SLMs (in terms of billions of parameters) are shown on the x-axis, whereas average AUROCs are shown on the y-axis. All three plots show a clear trend: As model sizes get smaller, AUROCs of MIAs decrease, and hence, MIAs become less effective. This observation suggests that smaller models, due to their reduced memorization capacity, exhibit fewer distinguishing characteristics between training and non-training samples, making MIAs more challenging in the context of SLMs. This key observation motivated us to propose a new MIA that is more effective specifically in the context of SLMs.

\vspace{-6pt}
\section{The Win-k Attack} \label{sec:WinK}
\vspace{-4pt}

\subsection{Attack Intuition and Explanation}

Motivated by the reduced performance of MIAs on SLMs, we propose a new MIA called \textbf{win-k}, which builds on top of the state-of-the-art \textbf{min-k} attack. We choose to build on top of min-k rather than the other attacks since min-k usually outperforms the other attacks. The original min-k attack takes the individual token-level log probabilities, sorts them in ascending order, and then selects the bottom k\% fraction to construct $f(x;M)$. In other words, it is a token-level approach. In contrast, win-k proposes to compute \textit{window-level} scores rather than token-level scores. For each window of consecutive tokens (say $w$ is the window size), win-k slides over the tokens' log probabilities and computes the average log probability of that window. Then, window-level scores are sorted in ascending order, and the bottom k\% fraction of the window-level scores are used to construct $f(x;M)$. Thus, win-k can identify if a \textit{window} of consecutive tokens collectively has a low log probability rather than focusing on single tokens. 

\begin{figure*}[!t]
\centering
    \includegraphics[width=.95\textwidth]{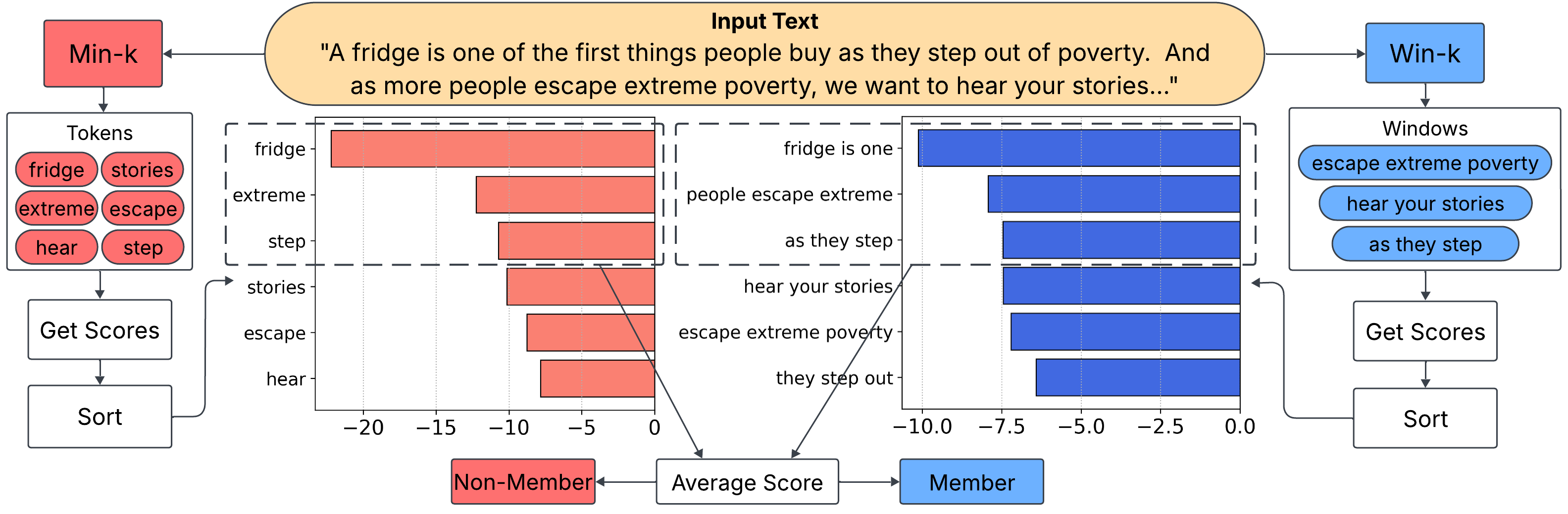} 
    \vspace{-8pt}
    \caption{Overview and comparison between min-k and win-k attacks.}
    \vspace{-8pt}
    \label{fig:wink-arch}
\end{figure*}

A visual overview and comparison between the behaviors of min-k and win-k can be found in Figure \ref{fig:wink-arch}. It can be observed that min-k decides that this sample is a non-member based on the average of the three tokens with the lowest log probability (fridge, extreme, step). Yet, win-k considers windows of three consecutive tokens (i.e., $w$ = 3) and decides that this sample is a member based on the average of the three windows with the lowest scores. We highlight that in min-k, the log probabilities vary greatly, e.g., between -22 and 0. However, in win-k, the range is smaller (e.g., between -10 and 0) and the scores have lower variance. Tokens with low log probabilities can still cause the overall window to have a lower score, e.g., ``fridge is one'' is the lowest-scoring window due to ``fridge'' having a low log probability. However, since the other two consecutive tokens in ``fridge is one'' balance the window's score, the average score is no longer very negatively impacted by ``fridge''. Thus, win-k can correctly predict this sample as a member, whereas min-k incorrectly predicts it as a non-member.  



\vspace{-4pt}
\subsection{Technical Description of Win-k}

Let $w$ be the window size parameter. For sample $x$, let $s_j$ denote the subsequence of tokens starting at $x_j$ and containing the next $w$ tokens, i.e.: $s_j = (x_j, x_{j+1}, ..., x_{j+w-1})$. We denote by $\text{logprob}(s_j)$:
\begin{equation} \label{eq:logprob}
    \text{logprob}(s_j) = \sum\limits_{i=j}^{i=j+w-1} \text{log}(Pr(x_i|x_{<i}))
\end{equation}
To eliminate the effect of $w$, $\text{logprob}(s_j)$ is normalized by $w$ to obtain the score of $s_j$, denoted by $\text{score}(s_j)$:
\begin{equation} \label{eq:score}
    \text{score}(s_j) = \frac{\text{logprob}(s_j)}{w}
\end{equation}
Given a sample $x$, win-k constructs all token subsequences $s_j$ from $x$, calculates their $\text{logprob}(s_j)$ and $\text{score}(s_j)$, sorts them in ascending order, and finds the bottom k\% of the scores. Finally, these bottom k\% scores are aggregated to arrive at the membership score of the whole sample $x$, i.e., $f(x;M)$. An algorithmic summary of the proposed win-k attack is shown in Algorithm \ref{alg:wink}.

\begin{algorithm}[!t]
\caption{Pseudocode of the win-k attack}
\DontPrintSemicolon
\label{alg:wink}
\SetKwInOut{Input}{Input}\SetKwInOut{Output}{Output}
\Input{Sample $x = (x_1,x_2,...,x_T)$, model $M$, window size $w$, fraction $k$}
\Output{Membership score of sample $x$, i.e., $f(x;M)$}
\BlankLine
Initialize an empty list: $\texttt{scoreList} \gets [~]$ \;
\For{$j = 1$ to $T-w+1$} {
Construct $s_j \gets (x_j, x_{j+1},...,x_{j+w-1}$) \;
Obtain $\text{logprob}(s_j)$ via Equation \ref{eq:logprob} using $M$ \;
Obtain $\text{score}(s_j)$ via Equation \ref{eq:score} \;
Append $\text{score}(s_j)$ to \texttt{scoreList} \;
}
Sort \texttt{scoreList} in ascending order \;
$\gamma \gets k \times T$ \;
\textbf{return} $\frac{1}{\gamma} \sum\limits_{i=1}^{\gamma} \texttt{scoreList}[i] $
\end{algorithm}

\vspace{-4pt}
\subsection{Why Does Win-k Work?}

\begin{figure*}[!t]
\centering
    \includegraphics[width=.995\textwidth]{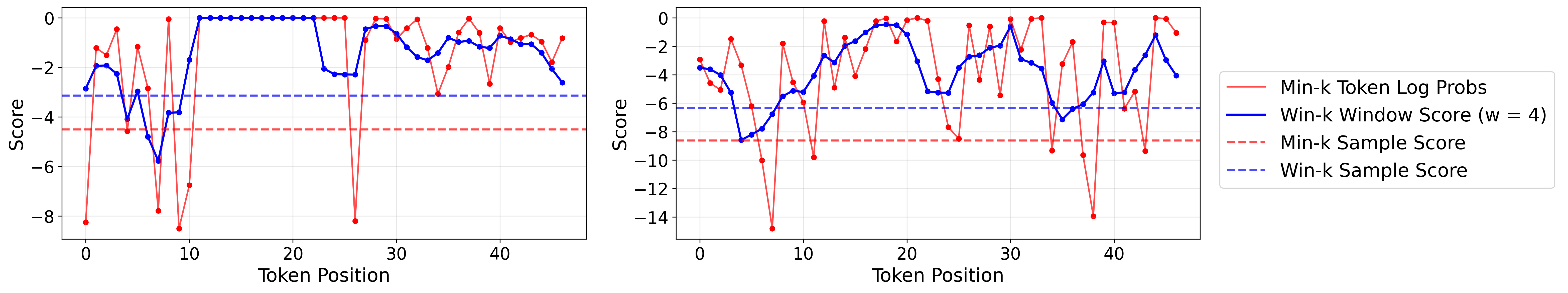}
    \vspace{-18pt}
    \caption{Scores produced by min-k and win-k for individual tokens and the whole sample. Member sample on the left, non-member sample on the right.}
    \vspace{-6pt}
    \label{fig:member-vs-nonmember}
\end{figure*}

An interesting question is why win-k works. To answer this question, we perform the following experiment. We randomly select one member sample and one non-member sample from the AGNews dataset, and obtain the scores produced for these samples by GPT-Neo 125M. The left plot in Figure \ref{fig:member-vs-nonmember} shows the results for the member sample, and the right plot shows the results for the non-member sample. Both plots contain four lines: (i) the log probabilities $\text{log}(Pr(x_i|x_{<i}))$ of individual tokens in the sample which are used by min-k, (ii) the window-level scores $\text{score}(s_j)$ of subsequences which are used by win-k, where $j \in [1,T-w]$, (iii) the aggregate min-k score for the whole sample shown by the red dashed line, and (iv) the aggregate win-k score for the whole sample shown by the blue dashed line. The fraction is $k$ = 30\%.

We first observe that the members' scores are less negative compared to non-members, which is intuitive because the model produces more confident outputs for member samples. However, an important difference between min-k's scores and win-k's scores is their variance. We can observe from the red curve (min-k) that the variance is quite high, especially in the case of non-members. In contrast, the blue curve (win-k) has lower variance and is more stable. Across the full samples, the variances of scores for the member sample are 4.72 in min-k and 1.21 in win-k; and the variances for the non-member sample are 10.25 in min-k and 2.24 in win-k. Indeed, variances of non-member samples are generally higher, and they are particularly much higher in min-k compared to win-k. 

SLMs have limited capacity, and their approximation of $Pr(x_i|x_{<i})$ can be noisy compared to LLMs. As a result, token-level log probabilities exhibit higher variance. This higher variance causes the membership score $f(x;M)$ in min-k to be dominated by the few tokens with strongly negative log probabilities. For example, even with $k$ = 30\%, we observe from Figure \ref{fig:member-vs-nonmember} that the dashed red lines are much lower than the average behavior of the individual tokens (i.e., average of the regular red lines). The fact that min-k is heavily affected by a few low-probability tokens, despite many tokens' probabilities being high, causes it to perform suboptimally. In contrast, the dashed blue lines (win-k) are closer to the average of the regular blue lines, i.e., average subsequence scores. Thus, we can conclude that the membership score $f(x;M)$ computed by win-k acts as a better representative of the whole sample compared to min-k. 



\section{Experiments and Discussion} \label{sec:Experiments}
\vspace{-4pt}

In this section, we provide an experimental evaluation of win-k using several SLMs, datasets, and hyperparameters. First, we describe our experiment setup. Second, we compare the effectiveness of win-k against other attacks. Third, we perform a hyperparameter analysis. Finally, we study the impacts of model training and data characteristics on attack effectiveness. 

\vspace{-4pt}
\subsection{Experiment Setup} \label{sec:ExperimentSetup}

\textbf{Models.} We perform experiments with three model families: GPT-Neo \cite{gpt-neo}, Pythia \cite{biderman2023pythia}, and MobileLLM \cite{liu2024mobilellm}. Since our work focuses on SLMs, we pick those models with $\leq$ 1.5B parameters. We use the following models in our experiments: GPT-Neo 125M; Pythia 70M, 160M, 410M, 1B; MobileLLM 125M, 350M, 600M. 


\textbf{Datasets.} We fine-tune SLMs on the following three datasets which are commonly used in the literature: WikiText, AGNews, and XSum. We created different versions of the datasets with different sample lengths: $T$ = 16, 32, 64, and 128. We use $T$ = 32 by default, but report results with varying $T$ in Section \ref{sec:training-params}. To test MIA effectiveness, we construct balanced test sets that contain 350 members (used in fine-tuning) and 350 non-members (not used in fine-tuning). 

\textbf{Fine-Tuning Parameters.} All models are fine-tuned using supervised fine-tuning (SFT) via the \textsc{SFTTrainer} framework. The maximum sequence length is set to 2048 tokens, number of epochs is set to 2 (experiments are done with varying numbers of epochs in Section \ref{sec:training-params}), batch size is set to 8, gradient accumulation is performed over 4 steps, and the learning rate is $3 \times 10^{-5}$. 

\textbf{Attack Hyperparameters.} We compare win-k against attacks presented in Section \ref{sec:MIA}. For the neighborhood attack, the number of neighbors per sample is 100 and BERT was used as the masked language model during neighbor generation. For min-k and win-k, we experiment with varying $k \in \{5\%, 10\%, 20\%, ..., 90\%\}$ and $w \in \{1, 2, 3, ..., 10\}$, and report the best results. We perform a hyperparameter analysis for win-k in Section \ref{sec:hyperparameters}.

\textbf{Evaluation Metrics.} The membership score $f(x;M)$ of a sample is compared against a threshold $\delta$ to predict whether that sample is a member or not. A true positive occurs when a member sample is predicted as a member, a false positive occurs when a non-member sample is predicted as a member, a true negative occurs when a non-member is predicted as a non-member, and a false negative occurs when a member is predicted as a non-member. Then, the effectiveness of MIAs are quantitatively evaluated using three metrics:
\vspace{-4pt}
\begin{itemize}
    \item \textbf{AUROC} (Area Under Receiver Operating Characteristics Curve): The ROC curve plots the TPR (True Positive Rate) versus the False Positive Rate (FPR) across different thresholds. AUROC is the area under the ROC curve. AUROC = 50\% indicates no discriminative power (i.e., random guessing), whereas AUROC = 100\% indicates perfect predictions. 
    \item \textbf{TPR @ 1\% FPR}: This metric measures the TPR of an attack under the stringent FPR requirement of FPR = 1\%.
    \item \textbf{FPR @ 99\% TPR}: This metric measures the FPR of an attack under the stringent TPR requirement of TPR = 99\%. 
\end{itemize}

\subsection{Comparison with Existing MIAs}

\begin{table}[!t]
\centering
\caption{AUROCs of different MIAs with varying models and datasets. MobLM is short for MobileLLM, Nbrhood is short for the Neighborhood attack, Lowercs is short for the Lowercase attack. The best attack in each case is highlighted in bold.}
\label{tab:auroc}
\setlength{\tabcolsep}{2pt}
\begin{tabular}{|c|c|cccccccc|}
\hline
\multirow{2}{*}{Dataset} & \multirow{2}{*}{Attack} & 
GPT-Neo & Pythia & Pythia & Pythia & Pythia & \scriptsize MobLM & \scriptsize  MobLM & \scriptsize MobLM \\
& & 125M & 70M & 160M & 410M & 1B & 125M &  350M & 600M \\
\hline
\renewcommand{\arraystretch}{1.2}
\multirow{6}{*}{\rotatebox{90}{\centering WikiText}} & Nbrhood & 67.0\% & 62.3\% & 61.8\% & 76.6\% & 83.6\% & 57.9\% & 59.9\% & 65.8\% \\
 & Lowercs & 66.7\% & 65.6\% & 66.3\% & 81.1\% & 90.2\% & 59.2\% & 59.2\% & 68.7\% \\
 & Loss & 74.4\% & 74.1\% & 77.1\% & 95.3\% & 98.5\% & 59.4\% & 60.2\% & 85.2\% \\
 & Zlib & 73.6\% & 73.6\% & 76.9\% & 95.1\% & 98.4\% & 59.7\% & 59.8\% & 83.7\% \\
 & Min-k & 76.0\% & 74.6\% & 77.6\% & \textbf{96.0\%} & \textbf{98.7\%} & 63.6\% & 63.9\% & \textbf{85.3\%} \\
 & Win-k & \textbf{76.3\%} & \textbf{75.1\%} & \textbf{78.9\%} & \textbf{96.0\%} & 98.4\% & \textbf{65.1\%} & \textbf{64.1\%} & 77.2\% \\ 
\hline
\multirow{6}{*}{\rotatebox{90}{\centering AGNews}} & Nbrhood & 78.3\% & 69.3\% & 68.4\% & 82.2\% & 91.0\% & 62.9\% & 65.3\% & 76.2\% \\
 & Lowercs & 80.1\% & 71.7\% & 71.2\% & 84.2\% & 96.9\% & 65.0\% & 67.2\% & 88.2\% \\
 &  Loss & 85.1\% & 80.5\% & 79.4\% & 90.7\% & 98.1\% & 68.6\% & 70.6\% & 94.0\% \\
 & Zlib & 83.6\% & 79.1\% & 78.0\% & 89.5\% & 97.4\% & 66.8\% & 69.1\% & 92.5\% \\
 & Min-k & 86.6\% & 81.2\% & 81.8\% & 92.9\% & 98.3\% & \textbf{76.8\%} & \textbf{81.2\%} & \textbf{94.2\%} \\
 & Win-k & \textbf{87.9\%} & \textbf{83.4\%} & \textbf{83.9\%} & \textbf{93.2\%} & \textbf{98.5\%} & 76.0\% & 79.4\% & 90.8\% \\ 
\hline
\multirow{6}{*}{\rotatebox{90}{XSum}} & Nbrhood & 63.2\% & 61.8\% & 67.6\% & 77.5\% & 86.3\% & 57.1\% & 59.4\% & 67.2\% \\
 & ~Lowercs~ & 64.7\% & 62.6\% & 68.3\% & 79.1\% & 89.6\% & 57.7\% & 60.2\% & 72.6\% \\
 & Loss & 68.7\% & 69.9\% & 77.1\% & 90.6\% & 95.8\% & 59.4\% & 62.0\% & 80.6\% \\
 & Zlib & 67.9\% & 69.1\% & 76.0\% & 89.5\% & 95.4\% & 59.1\% & 61.4\% & 78.1\% \\
 & Min-k & 69.2\% & 69.9\% & 77.2\% & \textbf{90.8\%} & \textbf{95.9\%} & 60.2\% & 63.9\% & \textbf{80.7\%} \\
 & Win-k & \textbf{69.9\%} & \textbf{70.4\%} & \textbf{78.0\%} & \textbf{90.8\%} & 95.5\% & \textbf{61.6\%} & \textbf{64.8\%} & 75.1\% \\ 
\hline
\end{tabular}
\vspace{-6pt}
\end{table}

In this section, we compare win-k with existing MIAs to demonstrate its superior effectiveness. Table \ref{tab:auroc} contains the AUROCs of different MIAs under 8 different models and 3 fine-tuning datasets. The attack with the highest AUROC (i.e., the most effective attack) in each case is highlighted in bold. When there is a tie, both attacks are highlighted in bold. 

In summary, Table \ref{tab:auroc} shows that win-k has the highest AUROC among all attacks in 17 out of 24 cases. Considering that AUROC measures the area under the ROC curve in general, this result demonstrates that win-k is generally more effective than the other attacks. We note that win-k outperforms the other MIAs more consistently especially when models are smaller, e.g., GPT-Neo 125M, Pythia 70M, and Pythia 160M. On larger models such as Pythia 410M or Pythia 1B, min-k can be tied with win-k or min-k can surpass win-k by a small amount. This shows that win-k is indeed better specifically for smaller language models. Another interesting observation is that win-k performs relatively worse on MobileLLM compared to GPT-Neo and Pythia families. A reason behind this could be the tokenizers. GPT-Neo and Pythia use similar tokenizers (GPT2Tokenizer and GPTNeoXTokenizer), which are both based on byte-pair encoding. Their vocabulary sizes are the same (50,257 tokens). Yet, MobileLLM uses a Llama-based tokenizer for which the vocabulary size is 32,000. This significant difference in tokenization may be causing min-k and win-k to perform similarly on MobileLLM, whereas win-k outperforms min-k on other models. 

\begin{table}[!t]
\centering
\caption{TPR @ 1\% FPR of different MIAs with varying models and datasets.}
\label{tab:tpr}
\setlength{\tabcolsep}{2pt}
\begin{tabular}{|c|c|cccccccc|}
\hline
\multirow{2}{*}{Dataset} & \multirow{2}{*}{Attack} & 
GPT-Neo & Pythia & Pythia & Pythia & Pythia & \scriptsize MobLM & \scriptsize  MobLM & \scriptsize MobLM \\
& & 125M & 70M & 160M & 410M & 1B & 125M &  350M & 600M \\
\hline
\renewcommand{\arraystretch}{1.2}
\multirow{6}{*}{\rotatebox{90}{\centering WikiText}} & Nbrhood & 4.3\% & 4.0\% & 3.1\% & 10.3\% & 12.0\% & 0.6\% & 1.7\% & 1.7\% \\
 & Lowercs & 4.6\% & 2.9\% & 5.1\% & 19.7\% & 55.7\% & 2.6\% & 2.6\% & 12.6\% \\
 & Loss & 3.1\% & 2.6\% & 5.1\% & 24.9\% & 69.1\% & 1.1\% & 1.4\% & 13.4\% \\
 & Zlib & 4.0\% & 4.0\% & 5.4\% & 27.7\% & 58.9\% & 0.9\% & 0.3\% & 13.4\% \\
 & Min-k & 4.6\% & 4.6\% & 5.7\% & 30.3\% & \textbf{75.4\%} & 4.6\% & 2.3\% & \textbf{15.4\%} \\
 & Win-k & \textbf{6.3\%} & \textbf{7.1\%} & \textbf{7.1\%} & \textbf{33.1\%} & 70.3\% & \textbf{5.1\%} & \textbf{4.0\%} & \textbf{15.4\%} \\ 
\hline
\multirow{6}{*}{\rotatebox{90}{\centering AGNews}} & Nbrhood & 1.7\% & 2.9\% & 2.3\% & 3.4\% & 23.7\% & 0.9\% & 1.1\% & 6.0\% \\
 & Lowercs & 6.0\% & 7.1\% & 5.4\% & 8.9\% & 27.4\% & 4.0\% & 5.1\% & 12.3\% \\
 & Loss & 2.3\% & 4.0\% & 1.7\% & 4.6\% & 38.9\% & 1.1\% & 1.4\% & 14.6\% \\
 & Zlib & 1.1\% & 1.1\% & 1.1\% & 1.4\% & 26.3\% & 1.1\% & 1.1\% & 2.0\% \\
 & Min-k & 8.6\% & 9.4\% & 8.0\% & 12.6\% & 40.3\% & 3.4\% & 7.1\% & 15.1\% \\
 & Win-k & \textbf{16.6\%} & \textbf{15.4\%} & \textbf{12.3\%} & \textbf{27.1\%} & \textbf{64.6\%} & \textbf{4.9\%} & \textbf{7.7\%} & \textbf{23.7\%} \\ 
\hline
\multirow{6}{*}{\rotatebox{90}{XSum}} & Nbrhood & 4.0\% & 3.1\% & 5.7\% & 8.6\% & 15.4\% & 2.6\% & 2.9\% & 6.0\% \\
 & ~Lowercs~ & \textbf{4.6\%} & \textbf{4.3\%} & 5.4\% & 4.3\% & \textbf{25.1\%} & 2.6\% & 4.0\% & \textbf{10.9\%} \\
 & Loss & 0.3\% & 1.4\% & 6.6\% & 6.0\% & 18.0\% & \textbf{4.6\%} & 2.9\% & 5.7\% \\
 & Zlib & 1.7\% & 1.1\% & 2.9\% & 8.0\% & 15.4\% & 2.0\% & 2.3\% & 7.1\% \\
 & Min-k & 0.6\% & 3.4\% & 5.7\% & \textbf{11.4\%} & \textbf{25.1\%} & 1.1\% & 3.7\% & 6.0\% \\
 & Win-k & 2.0\% & 2.3\% & \textbf{6.9\%} & 10.6\% & 17.7\% & 3.1\% & \textbf{4.3\%} & 7.1\% \\ 
\hline
\end{tabular}
\vspace{-4pt}
\end{table}

Next, we study the TPRs of the attacks @ 1\% FPR. The results in Table \ref{tab:tpr} show that win-k is the best performing attack in this metric in 17 out of 24 cases. Win-k particularly emerges as the best performer on WikiText and AGNews datasets. On the other hand, the Lowercase attack performs well on the XSum dataset. (If Lowercase did not exist, then win-k would have been the best-performing attack on XSum.) Based on all of our experiments, we observed that this exceptionally strong performance on Lowercase is limited to the strict setting of 1\% FPR, e.g., Lowercase does not perform as well in terms of other metrics or at other FPR thresholds. Third, we study the FPRs of the attacks @ 99\% TPR. As opposed to the other two metrics, lower values are better in this metric. Therefore, the lowest value in each case is highlighted in bold. We observe from Table \ref{tab:fpr} that win-k has the lowest FPRs in 12 out of 24 cases. Considering there are 6 attacks under comparison, win-k is still the best-performing attack. However, its superiority is not as significant in this metric compared to the other two metrics. To increase TPR, attacks typically choose the threshold that is more likely to predict samples as members. However, this approach mispredicts non-members as members, thereby increasing FPR. It is also worth noting that zlib, min-k, and loss attacks also perform reasonably well in terms of FPR @ 99\% TPR. Overall, based on our results, improving win-k to achieve lower FPR at high TPR regimes could be a good avenue for future work.

\begin{table}[!t]
\centering
\caption{FPR @ 99\% TPR of different MIAs with varying models and datasets.}
\label{tab:fpr}
\setlength{\tabcolsep}{2pt}
\begin{tabular}{|c|c|cccccccc|}
\hline
\multirow{2}{*}{Dataset} & \multirow{2}{*}{Attack} & 
GPT-Neo & Pythia & Pythia & Pythia & Pythia & \scriptsize MobLM & \scriptsize  MobLM & \scriptsize MobLM \\
& & 125M & 70M & 160M & 410M & 1B & 125M &  350M & 600M \\
\hline
\renewcommand{\arraystretch}{1.2}
\multirow{6}{*}{\rotatebox{90}{\centering WikiText}} & Nbrhood & 90.0\% & 95.1\% & 93.1\% & 85.4\% & 74.0\% & 98.0\% & 98.9\% & 97.1\% \\
 & Lowercs & 98.3\% & 96.9\% & 96.9\% & 96.0\% & 93.4\% & 97.4\% & 99.1\% & 99.7\% \\
 & Loss & 83.4\% & 90.9\% & 80.6\% & \textbf{32.9\%} & 10.0\% & 90.3\% & 96.3\% & \textbf{68.0\%} \\
 & Zlib & 90.9\% & \textbf{87.1\%} & \textbf{77.7\%} & 33.7\% & 12.9\% & 95.4\% & 97.7\% & 80.9\% \\
 & Min-k & \textbf{77.7\%} & 90.0\% & 83.4\% & 40.3\% & \textbf{8.6\%} & 88.9\% & 95.4\% & 86.3\% \\
 & Win-k & 80.9\% & 88.9\% & 79.7\% & 34.3\% & 17.1\% & \textbf{87.4\%} & \textbf{94.9\%} & 88.0\% \\ 
\hline
\multirow{6}{*}{\rotatebox{90}{\centering AGNews}} & Nbrhood & 79.1\% & 91.1\% & 88.3\% & 78.0\% & 48.0\% & 94.3\% & 90.6\% & 87.4\% \\
 & Lowercs & 73.7\% & 93.4\% & 88.6\% & 83.7\% & 25.4\% & 92.3\% & 88.3\% & 92.3\% \\
 & Loss & 61.4\% & 74.6\% & 78.6\% & 68.9\% & 6.3\% & 95.1\% & 88.6\% & \textbf{26.0\%} \\
 & Zlib & 69.1\% & 71.7\% & 79.4\% & 64.9\% & 6.9\% & 92.6\% & 94.3\% & 40.6\% \\
 & Min-k & \textbf{55.1\%} & 72.9\% & 73.7\% & \textbf{51.1\%} & 4.9\% & 77.1\% & \textbf{64.9\%} & 42.0\% \\
 & Win-k & 56.0\% & \textbf{68.3\%} & \textbf{66.6\%} & 52.0\% & \textbf{4.6\%} & \textbf{75.7\%} & 77.4\% & 55.4\% \\ 
\hline
\multirow{6}{*}{\rotatebox{90}{XSum}} & Nbrhood & 97.7\% & 98.3\% & 96.0\% & 90.3\% & 73.1\% & 98.3\% & 99.1\% & 97.4\% \\
 & ~Lowercs~ & 96.3\% & 98.0\% & 97.4\% & 87.1\% & 73.1\% & 99.1\% & 97.4\% & 97.1\% \\
 & Loss & 96.3\% & 93.1\% & 82.9\% & 56.0\% & 28.3\% & 94.6\% & 94.9\% & \textbf{84.0\%} \\
 & Zlib & 94.3\% & 93.4\% & 85.4\% & 59.4\% & 31.1\% & 96.0\% & 97.1\% & 89.7\% \\
 & Min-k & \textbf{89.1\%} & 92.6\% & 84.0\% & 51.1\% & 30.9\% & 95.1\% & \textbf{93.1\%} & 89.7\% \\
 & Win-k & 93.1\% & \textbf{88.3\%} & \textbf{80.0\%} & \textbf{50.6\%} & \textbf{25.1\%} & \textbf{93.1\%} & \textbf{93.1\%} & 86.3\% \\ 
\hline
\end{tabular}
\vspace{-6pt}
\end{table}

\vspace{-6pt}
\subsection{Analysis of Win-k Hyperparameters} \label{sec:hyperparameters}


There are two main hyperparameters in win-k: window size $w$ and fraction $k$. We report results with varying $w$ in Table \ref{tab:winsize}. Models in the table are listed from smallest to largest in terms of number of parameters (from top to bottom, 70M to 1B). Window sizes between $w$ = 2 and 10 are tested. For smaller models, e.g., less than 400M parameters, it can be observed that $w$ = 2, 3, or 4 yield better AUROC in many cases. For example, $w$ = 2 and 3 typically perform the best on AGNews, and $w$ = 3 and 4 typically perform the best on XSum. On WikiText, slightly higher $w$ such as $w$ = 4, 5, and 6 are better. On the other hand, for larger models such as MobileLLM 600M and Pythia 1B, larger $w$ are preferable. For example, on both WikiText and XSum datasets, $w$ = 8, 9, and 10 yield the highest AUROC. Overall, results show that the best $w$ is not fixed; it changes according to the size of the model. Smaller models tend to yield better results with $w$ values such as $w$ = 2, 3 or 4. In contrast, larger models yield better results with relatively larger $w$ such as $w$ = 8 or 9. We also observe that if $w$ is selected in parallel to this recommendation, the attack is not extremely sensitive to the precise value of $w$. It can be observed that the AUROCs in Table \ref{tab:winsize} vary by a moderate amount as $w$ changes. Thus, it is sufficient to choose a good enough $w$ following the above principle for win-k to perform well.

\begin{table}[!t]
\centering
\caption{Impact of $w$ on AUROCs of win-k under varying models and datasets.}
\label{tab:winsize}
\setlength{\tabcolsep}{2pt}
\scriptsize
\begin{tabular}{|c|c|c|c|c|c|c|c|c|c|c|}
\hline
Dataset & Model & 
$w$ = 2 & $w$ = 3 & $w$ = 4 & $w$ = 5 & $w$ = 6 & $w$ = 7 & $w$ = 8 & $w$ = 9 & $w$ = 10 \\
\hline
\renewcommand{\arraystretch}{1.2}
\multirow{7}{*}{\rotatebox{90}{\centering WikiText}} & \scriptsize Pythia 70M & 73.8\% & 74.1\% & 74.2\% & 74.3\% & 74.3\% & 74.2\% & 73.8\% & 73.7\% & 73.4\% \\
 & \scriptsize GPT-Neo 125M & 62.5\% & 62.8\% & 63.1\% & 63.1\% & 63.0\% & 62.9\% & 62.7\% & 62.4\% & 62.2\% \\
 & \scriptsize Pythia 160M & 77.4\% & 77.8\% & 77.9\% & 78.1\% & 78.2\% & 78.1\% & 77.8\% & 77.7\% & 77.5\% \\
 & \scriptsize MobLM 350M & 61.7\% & 61.8\% & 61.7\% & 61.7\% & 61.3\% & 61.3\% & 61.1\% & 61.0\% & 61.2\% \\
 & \scriptsize Pythia 410M & 95.0\% & 95.2\% & 95.4\% & 95.5\% & 95.6\% & 95.7\% & 95.8\% & 95.7\% & 95.7\% \\
 & \scriptsize MobLM 600M & 76.0\% & 76.1\% & 76.6\% & 77.0\% & 77.0\% & 77.4\% & 77.8\% & 78.1\% & 78.8\% \\
 & \scriptsize Pythia 1B & 98.0\% & 98.0\% & 98.1\% & 98.2\% & 98.3\% & 98.3\% & 98.3\% & 98.4\% & 98.3\% \\
\hline
\multirow{7}{*}{\rotatebox{90}{\centering AGNews}} & \scriptsize Pythia 70M & 82.3\% & 82.5\% & 82.1\% & 81.7\% & 81.2\% & 81.1\% & 80.9\% & 80.7\% & 80.6\%  \\
 & \scriptsize GPT-Neo 125M & 73.4\% & 73.4\% & 73.2\% & 73.0\% & 72.5\% & 72.0\% & 71.8\% & 71.5\% & 71.4\% \\
 & \scriptsize Pythia 160M & 82.2\% & 82.4\% & 82.2\% & 81.8\% & 81.4\% & 81.1\% & 81.0\% & 80.6\% & 80.3\% \\
 & \scriptsize MobLM 350M & 76.0\% & 75.6\% & 75.2\% & 74.8\% & 74.4\% & 74.0\% & 73.7\% & 73.4\% & 73.2\%  \\
 & \scriptsize Pythia 410M & 92.1\% & 92.1\% & 92.0\% & 91.9\% & 91.7\% & 91.5\% & 91.4\% & 91.3\% & 91.1\% \\
 & \scriptsize MobLM 600M & 87.7\% & 87.7\% & 87.9\% & 88.1\% & 88.1\% & 88.2\% & 88.4\% & 88.7\% & 89.2\% \\
 & \scriptsize Pythia 1B & 98.1\% & 98.2\% & 98.2\% & 98.3\% & 98.2\% & 98.3\% & 98.3\% & 98.2\% & 98.2\% \\
\hline
\multirow{7}{*}{\rotatebox{90}{XSum}} &   \scriptsize Pythia 70M & 69.6\% & 69.9\% & 69.9\% & 69.5\% & 69.2\% & 68.8\% & 68.9\% & 68.9\% & 68.9\% \\
 & \scriptsize GPT-Neo 125M & 60.3\% & 60.6\% & 60.6\% & 60.5\% & 60.5\% & 60.4\% & 60.2\% & 60.1\% & 60.1\% \\
 & \scriptsize Pythia 160M & 76.6\% & 77.1\% & 77.3\% & 77.2\% & 77.2\% & 77.0\% & 77.3\% & 77.1\% & 77.1\% \\
 & \scriptsize MobLM 350M & 63.3\% & 63.2\% & 63.1\% & 63.1\% & 62.9\% & 62.9\% & 62.9\% & 62.9\% & 63.0\% \\
 & \scriptsize Pythia 410M & 90.1\% & 90.4\% & 90.3\% & 90.2\% & 90.1\% & 90.0\% & 90.0\% & 89.9\% & 89.8\% \\
 & \scriptsize MobLM 600M & 74.2\% & 74.3\% & 74.5\% & 74.9\% & 74.9\% & 75.1\% & 75.3\% & 75.7\% & 76.1\% \\
 & \scriptsize Pythia 1B & 95.1\% & 95.2\% & 95.4\% & 95.4\% & 95.4\% & 95.4\% & 95.4\% & 95.4\% & 95.4\% \\
\hline
\end{tabular}
\end{table}

\begin{figure*}[!t]
\centering
    \includegraphics[width=.995\textwidth]{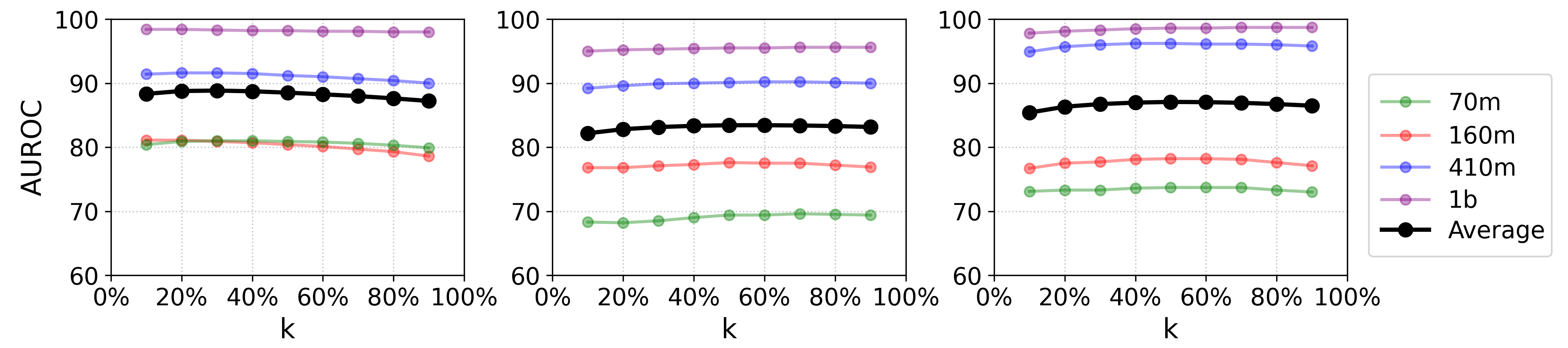}
    \vspace{-20pt}
    \caption{AUROCs of win-k with varying Pythia models and three datasets (left to right: AGNews, XSum, WikiText) under different $k$.}
    \vspace{-8pt}
    \label{fig:fraction-k}
\end{figure*}

In Figure \ref{fig:fraction-k}, we report results by varying the $k$ parameter. To improve statistical significance, we repeat the experiment with multiple Pythia models with varying sizes (70M, 160M, 410M, 1B) and all three datasets. The average AUROCs across all models are shown in the plots, in addition to the AUROCs of each individual model. We observe from the plots that $k$ values between 0.2 and 0.5 typically yield the highest AUROCs. Lower $k$, such as $k$ = 0.2 and 0.3, are better on AGNews (reducing $k$ to 0.1 yields lower AUROC). In contrast, $k$ = 0.4 and 0.5 work best on WikiText. $k$ = 0.4 works best on XSum as well; however, XSum shows the smallest change in AUROCs as the $k$ parameter changes. Overall, across all models and datasets, we observe the trend that $k$ should be selected neither too small nor too large. For example, $k$ = 0.1 and $k \geq 0.7$ often yield noticeable reductions in AUROC. To achieve the best results, we recommend $k$ between 0.3 and 0.5. 

\vspace{-6pt}
\subsection{Impact of Data and Fine-Tuning Related Parameters} \label{sec:training-params}

\begin{wrapfigure}{h}{0.43\textwidth}
\vspace{-30pt}
  \centering
  \includegraphics[width=0.4\textwidth]{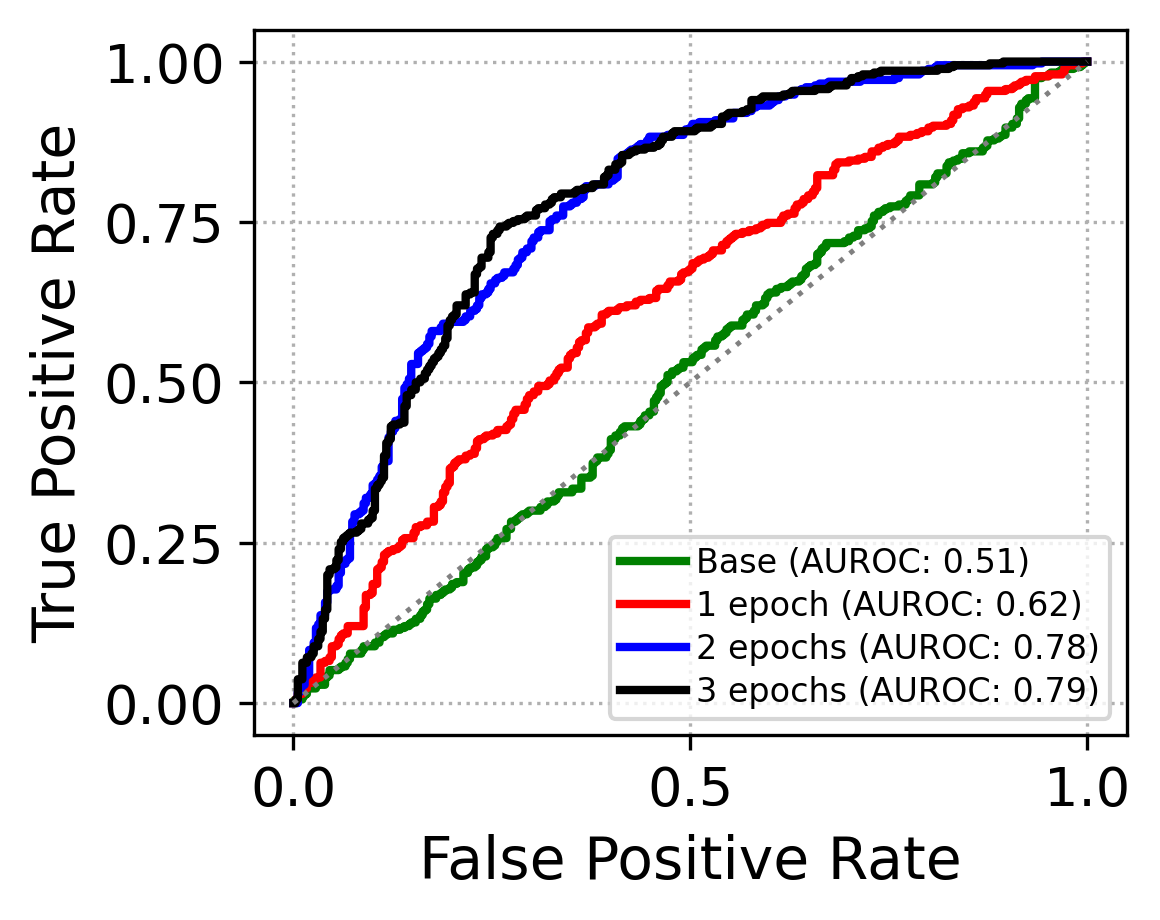} 
  \vspace{-8pt}
  \caption{Impact of changing number of epochs in terms of AUROC.}
  \label{fig:epochs}
  \vspace{-22pt}
\end{wrapfigure} 

Finally, we investigate the impacts of parameters related to data (text samples) and fine-tuning. In Figure \ref{fig:epochs}, we fine-tune the Pythia 160M model using the XSum dataset for varying numbers of epochs: 0 epochs (equal to the base model), 1 epoch, 2 epochs, and 3 epochs. It can be observed that executing win-k on the base model (0 epochs) indeed yields an AUROC close to 0.5, i.e., random guess. As we increase the number of epochs, AUROCs increase. There is a substantial increase from 0 epochs to 1 epoch, and also from 1 epoch to 2 epochs. However, the amount of increase from 2 epochs to 3 epochs is not very large, which shows that the model's susceptibility becomes saturated. Overall, it is intuitive that increasing the number of epochs increases model susceptibility, since the log probabilities produced by the model become more and more dominated by the fine-tuning dataset as the number of epochs is increased. It is important to note that a SLM like Pythia 160M becomes quickly vulnerable to our win-k attack, even with 1 or 2 epochs of fine-tuning.

\begin{table}[!t]
\centering
\caption{AUROCs of min-k and win-k with varying models and datasets under different number of tokens $T$.}
\label{tab:tokens}
\setlength{\tabcolsep}{6pt}
\begin{tabular}{|c|c|c|c|c|c|c|c|}
    \hline
    Dataset & Tokens & \multicolumn{2}{c|}{GPT-Neo 125M} & \multicolumn{2}{c|}{Pythia 160M} & \multicolumn{2}{c|}{Pythia 410M} \\
    \hline
    & & Min-k & Win-k & Min-k & Win-k & Min-k & Win-k  \\
    \hline
    \multirow{3}{*}{WikiText} & $T$ = 32 & 76.0\% & 71.7\% & 77.6\% & 78.9\% & 96.0\% & 96.0\% \\
    & $T$ = 64 & 83.1\% & 83.5\% & 84.8\% & 86.3\% & 98.8\% & 98.9\% \\
    & $T$ = 128 & 87.2\% & 87.0\% & 84.8\% & 86.3\% & 99.3\% & 99.5\% \\ \hline
    \multirow{3}{*}{XSum} & $T$ = 32 & 69.2\% & 69.9\% & 77.2\% & 78.0\% & 90.8\% & 90.8\%  \\
    & $T$ = 64 & 70.9\% & 71.9\% & 79.0\% & 80.0\% & 94.3\% & 94.8\% \\
    & $T$ = 128 & 76.9\% & 78.5\% & 84.2\% & 86.2\% & 97.6\% & 97.8\% \\ \hline
    \end{tabular}
    \vspace{-8pt}
\end{table}

In Table \ref{tab:tokens}, we investigate how the size of the text samples $x = (x_1,x_2,...,x_T)$ impacts attack effectiveness. For this experiment, we vary the value of $T$ by taking long samples (i.e., $T \geq$ 128) and truncating them to $T$ = 32, 64, and 128. We perform the experiment using three models (GPT-Neo 125M, Pythia 160M, and Pythia 410M), two datasets (WikiText and XSum), and two attacks (min-k and win-k). As the results in Table \ref{tab:tokens} show, increasing the $T$ typically yields a substantial increase in AUROC. The amount of increase is more noticeable in smaller models like GPT-Neo 125M and Pythia 160M since their AUROCs are smaller with $T$ = 32. In contrast, the AUROCs in Pythia 410M are already high when $T$ = 32; thus, the amount of increase from $T$ = 32 to 64 and 128 is less noticeable. Finally, it is worth noting that the results in Table \ref{tab:tokens} confirm that win-k usually continues to outperform min-k for varying $T$.

\vspace{-4pt}
\section{Related Work} \label{sec:RelatedWork}
\vspace{-2pt}

\textbf{Membership inference attacks (MIAs)} aim to determine whether a given sample was part of a model’s training dataset. First introduced in the context of classification models \cite{shokri2017membership}, MIAs have since been extended to a wide range of domains, including natural language processing (NLP) \cite{mireshghallah2022quantifying,shejwalkar2021membership}. More recently, MIAs have also been investigated in generative LLMs. Earlier attacks include the Loss attack \cite{carlini2019secret,yeom2018privacy} and the Lowercase and zlib attacks \cite{carlini2021extracting}. The Neighborhood attack was proposed in \cite{mattern2023membership} and the min-k attack was proposed in \cite{shi2024detecting}. Duan et al.~\cite{duan2024membership} developed the MIMIR benchmark to evaluate whether MIAs perform well on inferring membership of pre-training samples in LLMs.

Zhang et al.~\cite{zhang2024min} proposed an extension for min-k to better detect pre-training samples in LLMs. Non-biased MIAs (Nob-MIAs) were studied in \cite{eichler2024nob}. Nob-MIAs argue that inherent distributional biases exist between members and non-members in MIA evaluation, which can be addressed by creating non-biased and non-classifiable datasets for fairer MIA assessment. Document-level MIAs were proposed in \cite{meeus2024did}, in which the goal is to predict whether a large document (e.g., academic paper or book) was part of an LLM's training dataset. A similar, dataset-level membership inference task was studied in \cite{maini2024llm}. Finally, user inference attacks were proposed in \cite{kandpal2024user}, wherein an attacker infers whether a user's data was used for fine-tuning an LLM. These attacks are similar to MIAs, but they work at a much larger granularity (e.g., book, document, or user granularity rather than sample granularity). 





\textbf{Small language models (SLMs)} are language models with relatively modest parameter counts -- typically ranging from tens to a few hundred million parameters. They are designed to offer a balance between computational efficiency and task-specific performance. Unlike LLMs which require extensive computational resources, SLMs can be trained with limited hardware and are well-suited for edge deployment and on-device applications \cite{lu2024small,van2024survey}. Thus, the paradigm of fine-tuning pretrained SLMs on domain-specific and task-specific datasets is well-suited to improve downstream performance \cite{li2025efficient,phogat2024fine,xu2023fwdllm}. Popular SLM families include the Pythia, Qwen, MobileLLM, and SmolLM families \cite{biderman2023pythia,liu2024mobilellm,lu2024small}.

This paper is positioned at the intersection of MIAs and SLMs. Recent works show that LLMs are vulnerable to MIAs; however, these works utilize LLMs with billions of parameters, which are known to be prone to memorization and overfitting \cite{carlini2021extracting,carlini2022quantifying,hans2024like,kim2023propile}. In conrast, SLMs are much smaller. Furthermore, since SLMs are newly emerging, their vulnerability to MIAs is unknown. In this paper, we show that SLMs are also vulnerable to MIAs, but MIA effectiveness decreases as model size decreases. Motivated by this finding, we propose a new MIA, called win-k, which has higher effectiveness on SLMs compared to prior MIAs.

\vspace{-6pt}
\section{Conclusion} \label{sec:Conclusion}
\vspace{-2pt}

SLMs are rapidly gaining traction as efficient and deployable alternatives to LLMs, especially in resource-constrained and on-device AI applications. In this paper, we examined the vulnerability of SLMs to MIAs. Our empirical analysis revealed that the effectiveness of MIAs declines as model size decreases. We therefore proposed win-k, a new MIA which generalizes the min-k attack by aggregating log probability scores over sliding windows of tokens rather than individual tokens. Our experiments on eight SLMs across three datasets and three evaluation metrics showed that win-k outperforms prior attacks, especially on smaller models. Furthermore, we provided practical insights into the selection of win-k's hyperparameters and demonstrated its robustness across different training regimes and sample lengths. As future work, we plan to explore methods which can reduce the FPRs of win-k under strict TPR constraints and extend win-k to other modalities and tasks, e.g., multilingual models.


\vspace{-6pt}
\subsubsection{Acknowledgements} 

This study was supported by The Scientific and Technological Research Council of Turkiye (TUBITAK) under grants numbered 123E179 and 125E059. The authors thank TUBITAK for their support.
\color{black}

\bibliographystyle{splncs04}
\bibliography{references}

\end{document}